\documentclass{article}
\usepackage[numbers,sort,compress]{natbib}

\usepackage{times}
\usepackage{graphicx}
\usepackage{amsfonts, amsmath, amssymb}
\usepackage{multirow}
\usepackage{mathtools}
\usepackage{xcolor}

\title{Real-time Multi-Task Diffractive Deep Neural Networks via Hardware-Software Co-design}

\author
{Yingjie Li,$^{1}$ Ruiyang Chen,$^{1}$ Berardi Sensale Rodriguez,$^{1}$, Weilu Gao,$^{1}$ Cunxi Yu$^{1\ast}$\\
\\
\normalsize{$^{1}$Electrical and Computer Engineering Department, University of Utah}\\
\normalsize{50 S Central Campus Road, Salt Lake City, Utah, USA, 84112}\\
\\
\normalsize{$^\ast$E-mail:  cunxi.yu@utah.edu}
}

\date{}

\begin{document} 


\baselineskip24pt


\maketitle


\begin{abstract}
Deep neural networks (DNNs) have substantial computational requirements, which greatly limit their performance in resource-constrained environments. Recently, there are increasing efforts on optical neural networks and optical computing based DNNs hardware, which bring significant advantages for deep learning systems in terms of their power efficiency, parallelism and computational speed. Among them, free-space diffractive deep neural networks (D$^2$NNs) based on the light diffraction, feature millions of neurons in each layer interconnected with neurons in neighboring layers. However, due to the challenge of implementing reconfigurability, deploying different DNNs algorithms requires re-building and duplicating the physical diffractive systems, which significantly degrades the hardware efficiency in practical application scenarios. Thus, this work proposes a novel hardware-software co-design method that enables first-of-its-like \textbf{real-time} multi-task learning in D$^2$2NNs that automatically recognizes which task is being deployed in real-time. Our experimental results demonstrate significant improvements in versatility, hardware efficiency, and also demonstrate and quantify the robustness of proposed multi-task D$^2$NN architecture under wide noise ranges of all system components. In addition, we propose a domain-specific regularization algorithm for training the proposed multi-task architecture, which can be used to flexibly adjust the desired performance for each task. 
\end{abstract}


\section*{Introduction}
The past half-decade has seen unprecedented growth in machine learning with deep neural networks (DNNs). Use of DNNs represents the state-of-the-art in many applications, including large-scale computer vision, natural language processing, and data mining tasks \cite{lecun2015deep,silver2017mastering,senior2020improved}. DNNs have also impacted practical technologies such as web search, autonomous vehicles, and financial analysis \cite{lecun2015deep}. However, DNNs have substantial computational and memory requirements, which greatly limit their training and deployment in resource-constrained (e.g., computation, I/O, and memory bounded) environments. To address these challenges, there has been a significant trend in building high-performance DNNs hardware platforms. 
 {\color{black}While there has been significant progress in advancing customized silicon DNN hardware (ASICs and FPGAs) \cite{jouppi2017datacenter,senior2020improved} to improve computational throughput, scalability, and efficiency, their performance (speed and energy efficiency) are fundamentally limited by the underlying electronic components}. Even with the recent progress of integrated analog signal processors in accelerating DNNs systems which focus on accelerating matrix multiplication, such as Vector Matrix Multiplying module (VMM) \cite{ref_schlottmann2011}, mixed-mode Multiplying-Accumulating unit (MAC) \cite{ref_likamwa2016,ref_bankman2018, wang2018fully}, resistive random access memory (RRAM) based MAC \cite{ref_boybat2018,ref_jiang2018,ref_zand2018,wang2017memristors,hu2018memristor}, etc., the parallelization are still highly limited. Moreover, they are plagued by the same limitations of electronic components, with additional challenges in the manufacturing and implementation due to issues with device variability \cite{wang2017memristors,hu2018memristor}. 

Recently, there are increasing efforts on optical neural networks and optical computing based DNNs hardware, which bring significant advantages for machine learning systems in terms of their power efficiency, parallelism and computational speed \cite{silva2014performing,mengu2020scale,lin2018all,feldmann2019all,shen2017deep,tait2017neuromorphic,rahman2020ensemble,mengu2019analysis,luo2019design,hamerly2019large}. 
Among them, free-space \textit{diffractive deep neural networks} (D$^2$NNs) , which is based on the light diffraction, feature millions of neurons in each layer interconnected with neurons in neighboring layers. This ultrahigh density and parallelism make this system possess fast and high throughput computing capability.
Note that the diffractive propagations controlled by such physical parameters are differentiable, which means that such parameters can be optimized via conventional backpropagation algorithms \cite{mengu2020scale,lin2018all,mengu2019analysis} using \texttt{autograd} mechanism \cite{paszke2017automatic}.

In terms of hardware performance/complexity, one of the significant advantages of D$^2$NNs is that such a platform can be scaled up to millions of artificial neurons. In contrast, the design and DNNs deployment complexity on other optical architectures, e.g., integrated nantophotnics \cite{feldmann2019all,feldmann2020parallel} and silicon photnics \cite{tait2017neuromorphic}), can dramatically increase. For example, Lin et al. \cite{lin2018all} experimentally demonstrated various complex functions with an all-optical D$^2$NNs. In conventional DNNs, forward prorogation are computed by generating the feature representation with floating-point weights associated with each neural layer. In D$^2$NNs, such floating-point weights are encoded in the phase of each neuron of diffractive phase masks, which is acquired by and multiplied onto the light wavefunction as it propagates through the neuron. 
Similar to conventional DNNs, the final output class is predicted based on generating labels according to a given one-hot representation, e.g., the max operation over the output signals of the last diffractive layer observed by detectors. Recently, D$^2$NNs have been further optimized with advanced training algorithms, architectures, and energy efficiency aware training \cite{li2020class,mengu2019analysis,mengu2020scale}, e.g, class-specific differential detector mechanism improves the testing accuracy by 1-3\% \cite{li2020class}; \cite{mengu2020scale} improves the robustness of D$^2$NNs inference with data augmentation in training. 

However, due to the challenge of implementing reconfigurability in D$^2$NNs (e.g., 3D printed terahertz system \cite{lin2018all}), deploying a different DNNs algorithm requires re-building the entire D$^2$NNs system. In this manner, the hardware efficiency can be significantly degraded for multiple DNNs tasks, especially when those tasks are different but related. This has also been an important trend in conventional DNNs, which minimizes the total number of neurons and computations used for multiple related tasks to improve hardware efficiency, namely \textit{multi-task learning} \cite{ruder2017overview}. Note that, realizing different tasks directly from the input data features without separate inputs or user indications is challenging even in conventional DNNs system. In this work, we present the first-of-its-kind real-time multi-task D$^2$NNs architecture optimized in hardware-software co-design fashion, which enables sharing partial feature representations (physical layers) for multiple related prediction tasks. More importantly, \textbf{our system can automatically recognize which task is being deployed and generate corresponding predictions in real-time fashion, without any external inputs in addition to the input images.} Moreover, we demonstrate that the proposed hardware-software co-design approach is able to significantly reduce the complexity of the hardware by further reusing the detectors and maintain the robustness under multiple system noises. Finally, we propose an efficient domain-specific regularization algorithm for training multi-task D$^2$NNs, which offers flexible control to balance the prediction accuracy of each task (task accuracy trade-off) and prevent over-fitting. The experimental results demonstrate that our multi-task D$^2$NNs system can achieve the same accuracy for both tasks compared to the original D$^2$NNs, with more than 75\% improvements in hardware efficiency; and the proposed architecture is practically noise resilient under detector Gaussian noise and fabrication variations, where prediction performance degrades $\leq 1\%$ within the practical noise ranges.

\section*{Results and Discussion}\label{sec:results}

\begin{figure}[!htb]
  \includegraphics[width=1.\linewidth]{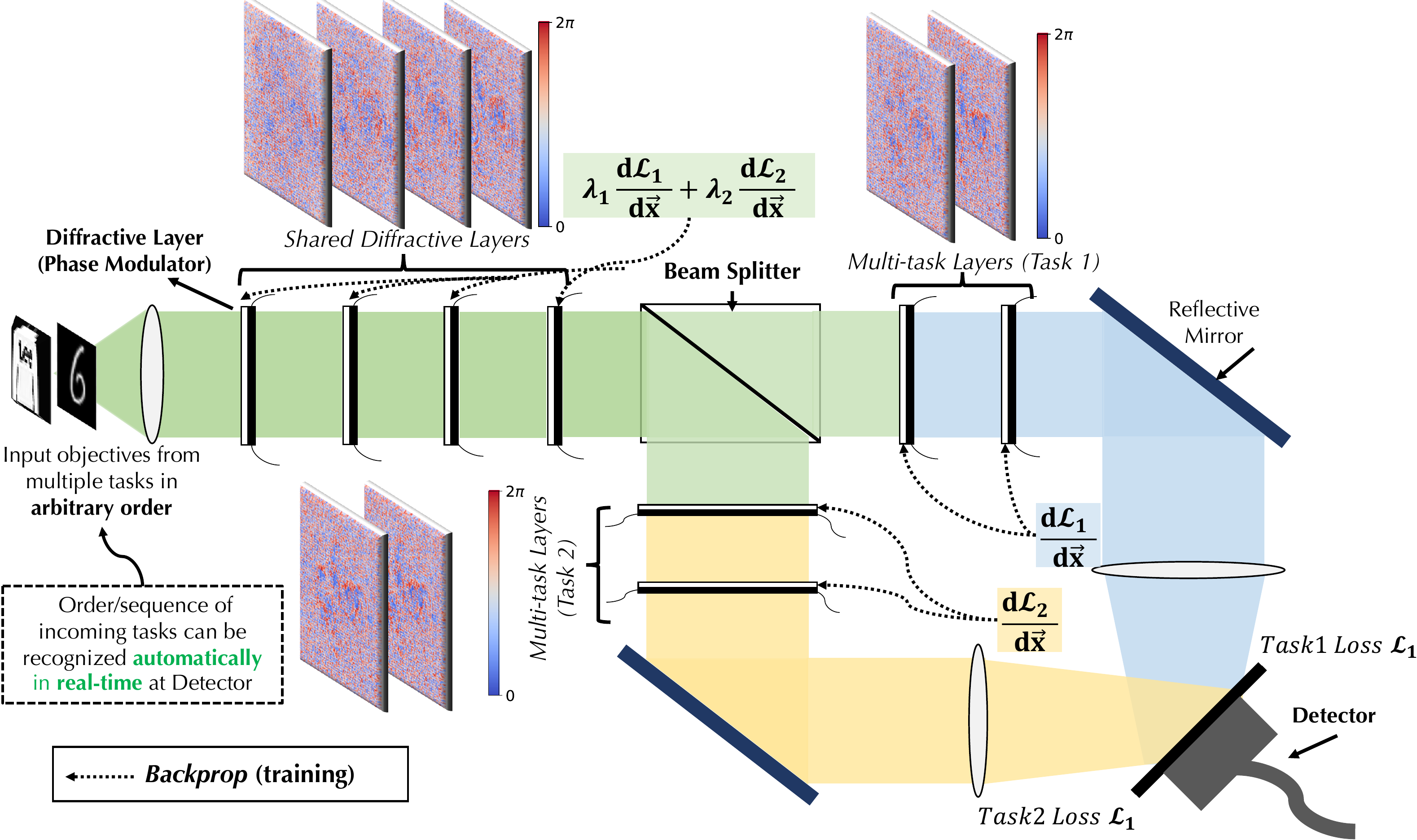}
\caption{{\bf Illustration of multi-task deep learning and multi-task D$^2$NN architecture with two image classification tasks deployed.} -- The proposed multi-task D$^2$NN architecture is formed by four shared diffractive layers and two multi-task layers, where the feed-forward computations have been re-used into multi-task layers using a beam splitter. With a novel training algorithm, the proposed architecture further reduces the hardware complexity that utilizes only ten detectors for both classification tasks, i.e., twenty different classes.}
\label{fig:architecture}
\end{figure} 

Figure \ref{fig:architecture} shows the proposed real-time multi-task diffractive deep neural network (D$^2$NN) architecture. Specifically, in this work, our multi-task D$^2$NN deploy image classification DNN algorithms with two tasks, i.e., classifying MNIST10 dataset and classifying Fashion-MNIST10 dataset. In a single-task D$^2$NN architecture for classification \cite{lin2018all}, the number of opto-electronic detectors positioned at the output of the system has to be equal to the number of classes in the target dataset. The predicted classes are generated similarly as conventional DNNs by selecting the index of the highest probability of the outputs (\texttt{argmax}), i.e., the highest energy value observed by detectors. Moreover, due to the lack of flexibility and reconfigurability of the D$^2$NN layers, deploying DNNs algorithms for $N$ tasks requires physically designing  $N$ D$^2$NN systems, which means $N$ times of the D$^2$NN layer fabrications and the use of detectors. Our main goal is to improve the cost efficiency of hardware systems while deploying multiple related ML tasks. Conceptually, the methodologies behind multi-task D$^2$NN architecture and conventional multi-task DNNs are the same, i.e., maximizing the shared knowledge or feature representations in the network between the related tasks \cite{ruder2017overview}.

Let the D$^2$NN multi-task learning problem over an input space $\mathcal{X}$, a collection of task spaces $\mathcal{Y}^n_{n\in[0,N]}$, and a large dataset including data points $\{x_i,y_i^1,...,y_i^N\}_I\in[D]$, where $N$ is the number of tasks and $D$ is the size of the dataset for each task. The hypothesis for D$^2$NN multi-task learning remains the same as conventional DNNs, which generally yields the following empirical minimization formulation:

\begin{equation}
    \min \limits_{\theta^{share}, {\theta^{1},\theta^{2},...,\theta^{N}}} ~~ \sum^{N}_{n=1} c^t \mathcal{L}(\theta^{share}, \theta^{t})
\end{equation}

where $\mathcal{L}$ is a loss function that evaluates the overall performance of all tasks. The finalized multi-task D$^2$NN will deploy the mapping, $f(x,\theta^{share},\theta^{n}) : \mathcal{X} \rightarrow \mathcal{Y}^n$, where $\theta^{share}$ are shared parameters in the \textit{shared diffractive layers} between tasks and task-specific parameters $\theta^{n}$ included in \textit{multi-task diffractive layers}. Specifically, in this work, we design and demonstrate the multi-task D$^2$NN with a two-task D$^2$NN architecture shown in Figure \ref{fig:architecture}. Note that the system includes four shared diffractive layers ($\theta^{share}$) and one multi-task diffractive layer for each of the two tasks. The multi-task mapping function becomes $f(x,\theta^{share},\theta^{1,2}) : \mathcal{X} \rightarrow \mathcal{Y}^2$, and can be then decomposed into:

\begin{equation}
   f(x,\theta^{share},\theta^{1,2}) = det(f^{1}(\frac{1}{2} \cdot f^{share}(x,\theta^{share}),\theta^{1}) + f^{2}(\frac{1}{2} \cdot f^{share}(x,\theta^{share}),\theta^{2}))\\
\end{equation}
\begin{equation}
   f^{share}: \mathcal{X} \rightarrow (\Re + \Im)^{\in 200\times200}, ~~ f^{1}, f^2: (\Re + \Im)^{\in 200\times200} \rightarrow (\Re + \Im)^{\in 200\times200}
\end{equation}

where $f^{share}, f^{1},$ and $f^2$ produce mappings in complex number domain that represent light propagation in phase modulated photonics. Specifically, the forward functionality of each diffractive layer and its dimensionality $\mathbb{R}^{200 \times 200}$ remains the same as \cite{lin2018all}. The output $det \in \mathbb{R}^{C \times 1}$ are the readings from $C$ detectors, where $C$ is the largest number of classes among all tasks; for example, $C=10$ for MNIST and Fashion-MNIST. The proposed multi-task D$^2$NN system is constructed by designing six phase modulators based on the optimized phase parameters in the four shared and two multi-task layers (Figure \ref{fig:architecture}), i.e., $\theta^{share},\theta^{1,2}$. The phase parameters are optimized with \textit{backpropogation} with gradient chain-rule applied on each phase modulation and adaptive momentum stochastic gradient descent algorithm (\texttt{Adam}). The design of phase modulators can be done with 3D printing or lithography to form a passive optical network that performs inference as the input light diffracts from the input plane to the output. Alternatively, such diffractive layer models can also be implemented with spatial light modulators (SLMs), which offers the flexibility of reconfiguring the layers with the cost of limiting the throughput and increase of power consumption.

Table \ref{tbl:comparisons} presents the performance evaluation and comparisons of the proposed architecture with other options of classifying both MNIST and Fashion-MNIST tasks. We compare our architecture with -- 1) singe-task D$^2$NN architecture, which requires two stand-alone D$^2$NN systems; 2) multi-task D$^2$NN architecture with the same diffractive architecture as Figure \ref{fig:architecture} but with two separate detectors for reading and generating the classification results. {\color{black}Specifically, we utilize \textit{Accuracy-Hardware product} (a.k.a. Acc-HW) metric. Regarding the hardware cost, we estimate the cost of the baseline and the proposed systems using the number of detectors. This is because the major cost of the system comes from detectors in practice and the cost of 3D-printed masks is negligible compared to detector cost. To evaluate the hardware efficiency improvements, we set single-task Acc-HW as the baseline, and the improvements of the multi-task D$^2$NN architectures using Equation \ref{eq:acc-hw}. We can see that our multi-task D$^2$NN architecture gains 75\% efficiency for MNIST task and 72\% for Fashion-MNIST task, by introducing a novel multi-task algorithm and modeling that detects 20 different classes (two sets) using only 10 detectors; and gains over 55\% and 50\% compared to using an architecture that requires two separate sets of detectors. }

\begin{equation}
\text{Acc-HW Product} = 1 \cdot \frac{Acc_{multi}}{Acc_{single}} \cdot \frac{HWCost_{multi}}{HWCost_{single}}; HWCost = \# Detectors.
\label{eq:acc-hw}
\end{equation}

\begin{table}[t]
\color{black}
\centering
\caption{{\bf Hardware efficiency comparison between single-task and multi-task D$^2$NN architectures.} For the multi-task D$^2$NN comparison, we compare the hardware efficiency and prediction accuracy between a dual-detection (20 detector regions) architecture and single-detection (10 detector regions). The efficiencies of different D$^2$NN architectures for MNIST and Fashion-MNIST tasks are evaluated using \textit{Accuracy-Hardware product} (a.k.a. Acc-HW), where hardware cost is estimated using the number of detectors.}
\label{tbl:comparisons}
\small
\begin{tabular}{|l|l|l|l|l|l|l|}
\hline
\multirow{2}{*}{\textbf{\begin{tabular}[c]{@{}l@{}}\end{tabular}}} & \multicolumn{2}{c|}{\textbf{Single-task system}} & \multicolumn{2}{c|}{\textbf{\begin{tabular}[c]{@{}c@{}}Multi-task system \\ w 10 Det-Regions\end{tabular}}} & \multicolumn{2}{c|}{\textbf{\begin{tabular}[c]{@{}c@{}}Multi-task system \\ w 20 Det-Regions\end{tabular}}} \\ \cline{2-7} 
 & \multicolumn{1}{c|}{\textit{MINST}} & \multicolumn{1}{c|}{\textit{F-MINST}} & \multicolumn{1}{c|}{\textit{MINST}} & \multicolumn{1}{c|}{\textit{F-MINST}} & \multicolumn{1}{c|}{\textit{MINST}} & \multicolumn{1}{c|}{\textit{F-MINST}} \\ \hline
Diffractive Layer Cost & 6$\times$200$\times$200 & 6$\times$200$\times$200 & \multicolumn{2}{l|}{(4+2+2)$\times$200$\times$200} & \multicolumn{2}{l|}{(4+2+2)$\times$200$\times$200} \\ \hline
Detector Cost & 10 & 10 & \multicolumn{2}{c|}{10} & \multicolumn{2}{c|}{10+10} \\ \hline
Accuracy & 0.981 & 0.889 & 0.977 & 0.886 & 0.979 & 0.883 \\ \hline
\textbf{Acc-HW Product} & 1 & 1 & \textbf{1.99}$\mathbf{\times}$ & \textbf{1.99}$\mathbf{\times}$ & {$\sim 1$}${\times}$ & {$\sim 1$}${\times}$ \\ \hline
\end{tabular}
\end{table}

   \begin{figure}[!htb]
      \includegraphics[width=1.\linewidth]{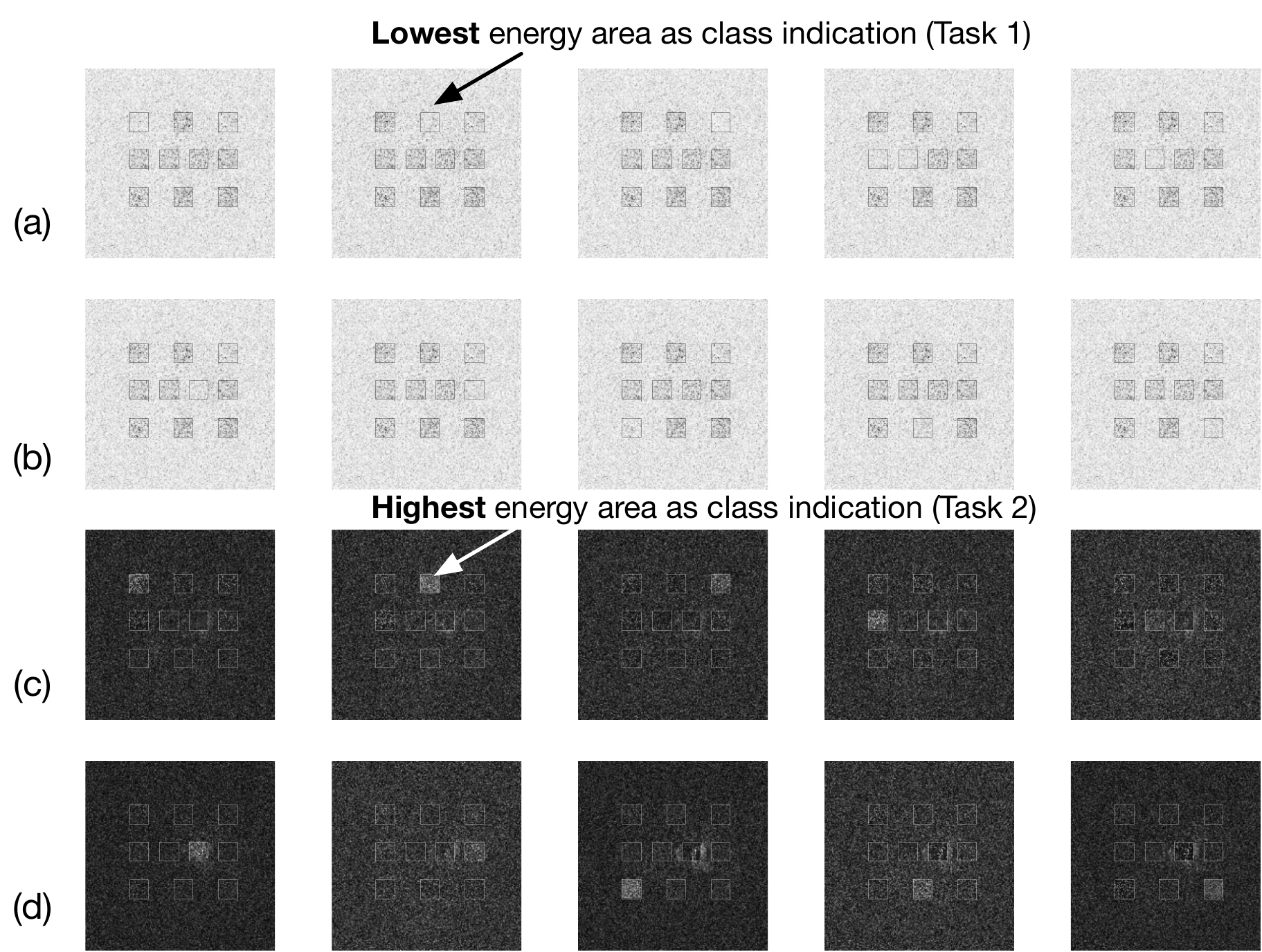}
        \caption{{\bf Modeling of ten classes for two different datasets with ten detectors.} (a) -- (b) One-hot encoding for classes 0 -- 9 of the first task (MNIST) represented using the energy value observed at the detectors. Final classes are produced using the index of the lowest energy area, i.e., \texttt{argmin}($det$). (c) -- (d) One-hot encoding for classes 0 -- 9 represented of the second task (e.g., Fashion-MNIST) using the energy value observed at the detectors. Final classes are produced using the index of the highest energy area, i.e., \texttt{argmax}($det$).}
        \label{fig:labeling}
    \end{figure}  
   
Figure \ref{fig:labeling} illustrates the proposed approach for producing the classes, which re-use the detectors for two different tasks. Specifically, for the multi-task D$^2$NN evaluated in this work, both MNIST and Fashion-MNIST have ten classes. Thus,   all the detectors used for one class can be fully re-utilized for the other. To enable an efficient training process, we use one-hot encodings for representing the classes similarly as the conventional multi-class classification ML models. The novel modeling introduced in this work that enables re-using the detectors is -- \textit{defining "1" differently in the one-hot representations}. As shown in Figure \ref{fig:labeling}(a)--(b), for the first task MNIST, the one-hot encoding for classes 0 -- 9 are presented, where each bounding box includes energy values observed at the detectors. In which case, "1" in the one-hot encoding is defined as the lowest energy area, such that the label can be generated as \texttt{argmin}($det$) -- the index of the lowest energy area. Similarly, Figure \ref{fig:labeling}(c)--(d) are the one-hot encodings for classes 0 -- 9 of the second task Fashion-MNIST, where label is the index of the highest energy area, i.e., \texttt{argmax}($det$). Therefore, ten detectors can be used to generate the final outputs for two different tasks that share the same number of classes, to gain extra 55\% and 50\% hardware efficiency of the proposed multi-task D$^2$NN (see Table \ref{tbl:comparisons}).

\begin{figure}
      \includegraphics[width=1.\linewidth]{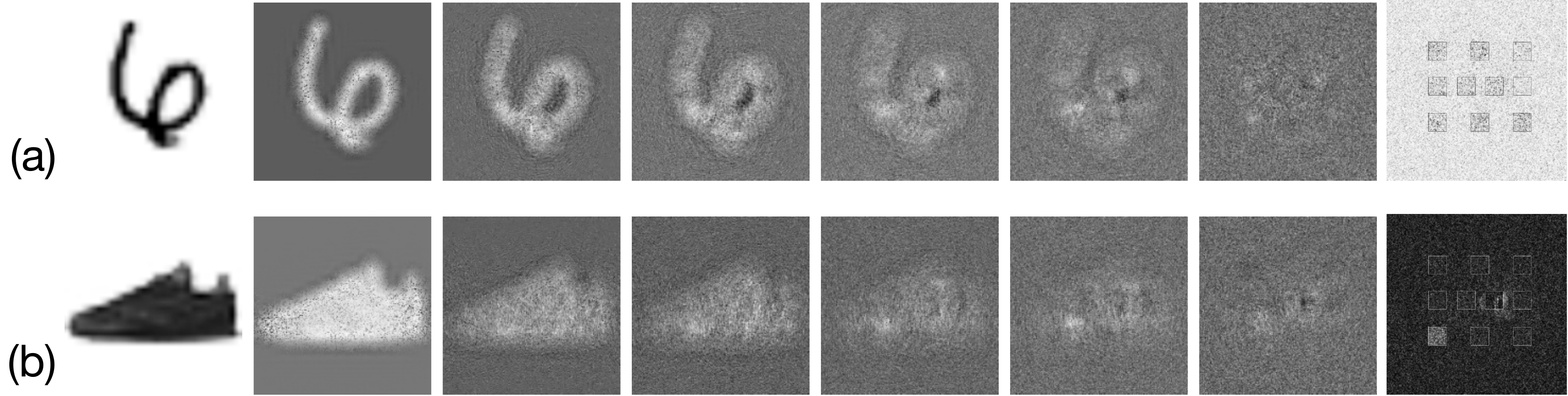}
        \caption{{\bf Visualization of propagations through multi-task D$^2$NN and the results on the detectors.} (a) Forward visualization of classifying MNIST10 sample with class=6, where the 7th detector has the lowest energy value. (b) Forward visualization of classifying Fashion-MNIST sample with class=7, where the 8th detector has the lowest energy value.}
        \label{fig:propagation}
\end{figure}  
    
Figure \ref{fig:propagation} includes visualizations of light propagations through multi-task D$^2$NN and the results on the detectors, where the input, internal results after each layer, and output are ordered from left to right. Figure  \ref{fig:propagation}(a) shows one example for classifying MNIST sample, where the output class is correctly predicted (class 7) by returning the index of the lowest energy detector. Figure \ref{fig:propagation}(a) presents an example for classifying Fashion-MNIST sample, where the output class is correctly predicted (class 8) by returning the index of the highest energy detector.

   \begin{figure}[!htb]
      \includegraphics[width=1.\linewidth]{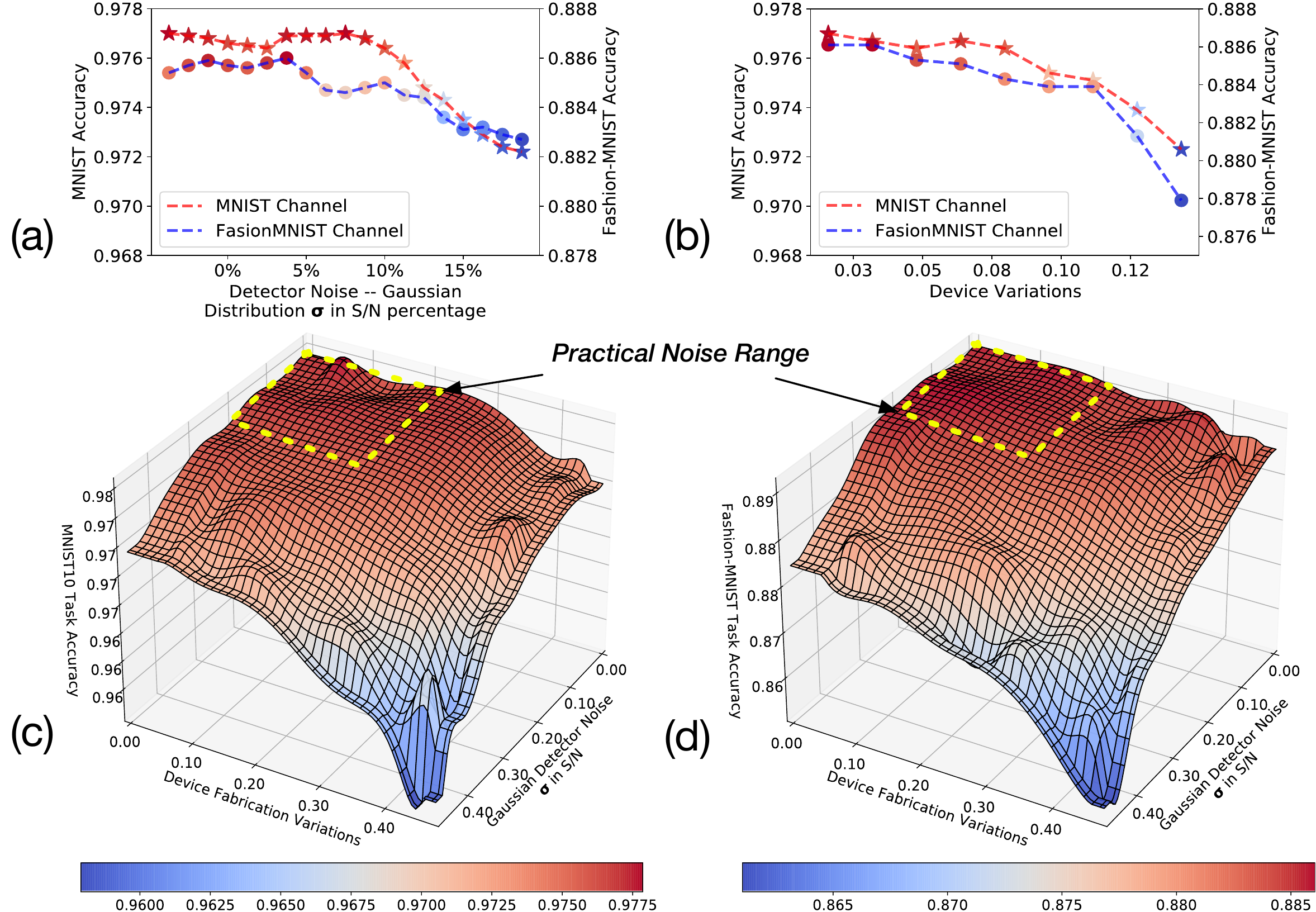}
            \caption{{\bf Evaluations of robustness against system noise of the proposed multi-task D$^2$NN, by considering a wide range of Gaussian noise in detectors and device variations in phase modulators.} Details of noise modeling in the proposed systems are discussed in Section Methods (Equations \ref{eq:det_noise} -- \ref{eq:combined_noise}). (a) Prediction performance evaluation under Gaussian detector noise with $\sigma$ shown in $S/N$ (Signal to Noise) $\in [0, 0.2]$. (b) Prediction performance evaluation under Gaussian device variations. (c) Evaluations of MNIST task accuracy under combined detector noise and device variations. (d) Evaluations of Fashion-MNIST task accuracy under combined detector noise and device variations.}
    \label{fig:noise_eval}
    \end{figure}

While building conventional multi-task DNN, it is well known that the robustness of the multi-task DNNs degrades compared to single-task DNNs, for each individual task. Such concerns become more critical in the proposed multi-task D$^2$NN system due to the potential system noise introduced by the fabrication variations, device variations, detector noise, etc. Thus, we comprehensively evaluate the noise impacts for our proposed multi-task D$^2$NN, by considering a wide range of Gaussian noise in detectors and device variations in phase modulators. Details of noise modeling in the proposed systems are discussed in Section Methods (Equations \ref{eq:det_noise} -- \ref{eq:combined_noise}). Figure \ref{fig:noise_eval} includes four sets of experimental results for evaluating the robustness of our system under system noise. Specifically, Figure \ref{fig:noise_eval}(a) evaluates the prediction performance of both tasks under detector noise, where the x-axis shows the $\sigma$ of a Gaussian noise vector $S/N$ (Signal to Noise), and the y-axis shows the accuracy. Figure \ref{fig:noise_eval}(b) evaluates the accuracy impacts from device variations of phase modulators, where the x-axis shows the phase variations of each optical neuron in the diffractive layer (note that phase value is $\in [0,2\pi]$), and the y-axis shows the accuracy. In practice, detector noise is mostly within 5\%, and device variations are mostly up to 0.2 (80\% yield). We can see that the prediction performance of the proposed system is resilient to a realistic noise range while considering only one type of noise. Moreover, in Figures \ref{fig:noise_eval}(c)--(d), we evaluate the noise impacts for MNIST and Fashion-MNIST, respectively, under both detector noise and device variations. While the accuracy degradations are much more noticeable when both noises become significantly, we observe that the overall performance degradations remain $\leq 1\%$ within the practical noise ranges. In summary, the proposed architecture is practically noise resilient.

\begin{figure}[!htb]
       \centering
      \includegraphics[width=1\linewidth]{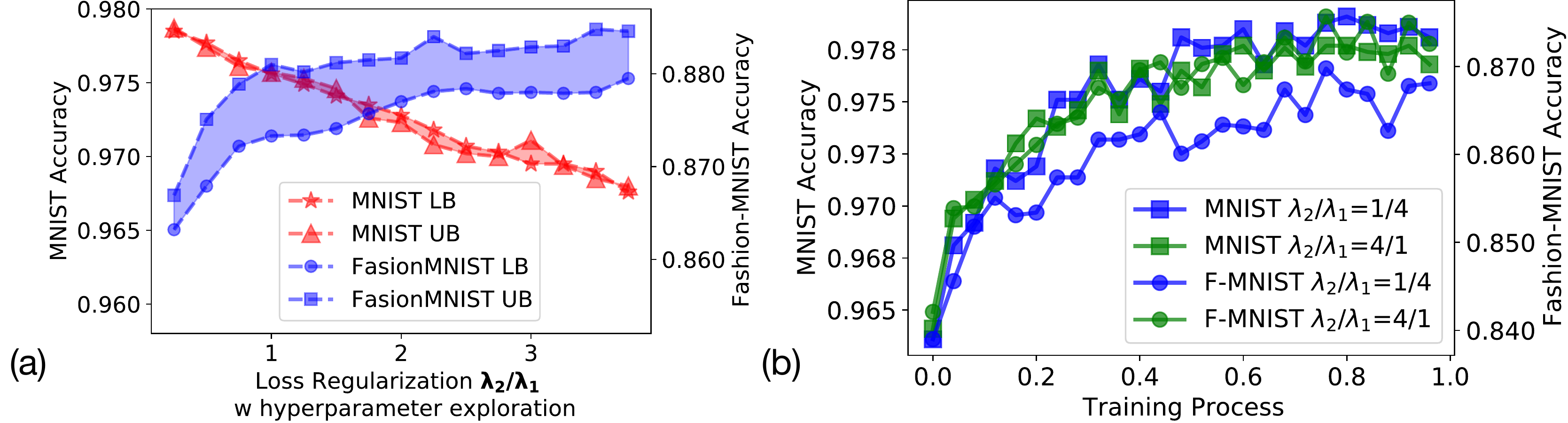}
            \caption{{\bf Evaluation of loss regularization for adjusting the performance of each task.} (a) Testing accuracy with different regularization factors. As $\frac{\lambda_2}{\lambda_1}$ increases (decreases), the final performance of the multi-task D$^2$NN will be bias to Fashion-MNIST (MNIST). We include results of 100 different hyperparameters for training. (b) Testing accuracy of both tasks during the training phase, where we can see that even the largest and smallest regularization factors do not cause overfitting.}
    \label{fig:algorithm_result}
\end{figure}  	

In multi-task learning, it is often needed to adjust the weight or importance of different prediction tasks according to the application scenarios. For example, one task could be required to have the highest possible prediction performance while the performance of other tasks are secondary. To enable such biased multi-task learning, the shared representations $\theta^{share}$ need to carefully adjusted. Figure \ref{fig:algorithm_result} demonstrates the ability to enable such biased multi-task learning using loss regularization techniques. Specifically, we propose to adjust the performance of different tasks using a novel domain-specific regularization function shown in Equation \ref{eq:loss_methods}, where $\lambda_1$ and $\lambda_2$ are used to adjust the task importance, with a modified \textit{L2 normalization} applied on multi-task layers only. The results with 100 trials of training (with different random seeds for initialization and slightly adjusted learning rate) are included in Figure \ref{fig:algorithm_result}(a). We can see that loss regularization is sufficient to enable biased multi-task learning in the proposed multi-task D$^2$NN architecture, regardless of the initialization and training setups. Moreover, Figure \ref{fig:algorithm_result}(b) empirically demonstrates that with even with very large or small regularization factors, the proposed loss regularization will unlikely overfit either of the tasks because of the adjusted L2 norm used in the loss function (Equation \ref{eq:loss}). Note that the adjusted L2 normalization only affects the gradients for $\theta^{1}$ and $\theta^{2}$, where $\lambda_{L2}$ is the weight of this L2 normalization.

\begin{equation}
    \mathcal{L}(\theta^{share}, \theta^{1,2}) = \underbrace{\lambda_1}_{\text{t1 factor}} \mathcal{L}_1(\theta^{share}, \theta^{1}) + \underbrace{\lambda_2}_{\text{t2 factor}} \cdot \mathcal{L}_2(\theta^{share}, \theta^{2}) + \underbrace{\lambda_{L2} \frac{\lambda_2}{\lambda_1} \cdot ( ({\theta^{1}})^2 + ({\theta^{2}})^2)}_{\text{adjusted L2 norm}}
    \label{eq:loss_methods}
\end{equation}

\section*{Methods}

\paragraph{Multi-task D$^2$NN Architecture} Figure \ref{fig:architecture} shows the design of the multi-task D$^2$NN architecture. Based on the phase parameters $\theta^{share}, \theta^1$, and $\theta^2$, there several options to implement the diffractive layers to build the multi-task D$^2$NN system. {For example, the passive diffractive layers can be manufactured using 3D printing for long-wavelength light (e.g. terahertz) or lithography for short-wavelength light (e.g. near-infrared), and active reconfigurable ones can be implemented using spatial light modulators. A 50-50 \textit{beam splitter} is used to split the output beam from the last shared diffractive layer into two ideally identical channels for multi-task layers. Coherent light source, such as laser diodes, is use in this system. At the output of two multi-task layers, the electromagnetic vector fields are added together on the detector plane. The generated photocurrent corresponding to the optical intensity of summed vector fields is measured and observed as output labels.} {\color{black}Regarding the real-time capability of the proposed system, the proposed architecture performs the same the system proposed in \cite{lin2018all}, where computation is executed at the speed of light and the information is processed on each neuron/pixel of the phase mask is highly parallel. Thus, the time of light flight is negligible and the determination factor for system hardware performance is dependent on the performance of THz detectors. For a detector with operation bandwidth $f$, the corresponding latency is $1/f$ and the largest throughput is $f$ frames/s/task. The minimum power requirement for this system is determined by the number of detector, $NEP$ (noise-equivalent-power), and , if we assume the loss and energy consumption associated with phase masks is negligible. In practice, considering a room-temperature VDI detector \footnote{https://www.vadiodes.com/en/products/detectors?id=214} operating at $\sim0.3~THz$ , $f=\sim 40~GHz$, and $NEP=2.1 pW/\sqrt[2]{Hz}$, the latency of the system will be 25 $ps$, throughput is $4 \times10^{10}$ $fps/task$ (frame/second/task), with power consumption 0.42 $uW$. In addition to mitigate the large cost of detectors, alternative materials can be used, such as graphene. For example, the specific detector performance shown in \cite{castilla2019fast} is $NEP=\sim80 pW/\sqrt[2]{Hz}$, and $f=\sim 300~MHz$. In which case, the system atency is $\sim30 ns$, such that the throughput is $3 \times 10^{8}$ $fps/task$ with the estimated minimum power 1.4 $uW$.}

\paragraph{Training and Inference of Multi-task D$^2$NN}  The proposed system has been implemented and evaluated using Python (v3.7.6) and Pytorch (v1.6.0). The basic components in the multi-task D$^2$NN PyTorch implementation includes 1) diffractive layer initialization and forward function, 2) beam splitter forward function, 3) detector reading, and 4) final predicted class calculation. First, each layer is composed of one diffractive layer that performs the same phase modulation as  \cite{lin2018all}. To enable high-performance training and inference on GPU core, we utilize for complex-to-complex Discrete Fourier Transform in PyTorch (\texttt{torch.fft}) and its inversion (\texttt{torch.ifft}) to mathematically model the same modulation process as \cite{lin2018all}. Beam splitter that evenly splits the light into \textit{transmitted light} and \textit{reflected light} is modeled as dividing the complex tensor produced by the shared layers in half. The trainable parameters are the phase parameters in the diffractive layers that modulate the incoming light. While all the forward function components are differentiable, the phase parameters can be simply optimized using automatic differentiation gradient mechanism (autograd). The detector has ten regions and each detector returns the sum of all the pixels observed (Figure \ref{fig:labeling}). To enable training with two different one-hot representations that allow the system to reuse ten detectors for twenty classes, the loss function is constructed as follows:

\begin{equation}
\begin{multlined}
\small
\mathcal{L}=\lambda_1 \cdot \underbrace{\texttt{MSELoss}(LogSoftmax(f(\theta^{share}, \theta^1, \mathcal{X}^1)~, ~(label^1 + 1)\%2~~)}_{\text{one-hot encoding with one "0" and nine "1s"}} \\ ~+~ \lambda_2 \cdot \underbrace{\texttt{MSELoss}(LogSoftmax(f(\theta^{share}, \theta^2, \mathcal{X}^2)~, ~label^2~)}_{\text{one-hot encoding with one "1" and nine "0s"}} \\
+ \underbrace{\lambda_{L2} \cdot \frac{\lambda_2}{\lambda_2} \texttt{L2}(\theta^1,\theta^2)}_{\text{L2 norm on multi-task diffractive layers}} \\
\label{eq:loss}
\end{multlined}
\end{equation}

The original labels $label^{1}$ and $label^2$ are represented in conventional one-hot encoding, i.e., one "1" with nine of "0s", and $label^{1}$ has been converted into an one-hot encoding with one "0" and nine "1s". Note that LogSoftmax function is only used for training the network, and the final predicted classes of the system are produced based on the values obtained at the detectors. With loss function shown in Equation \ref{eq:loss} and the modified one-hot labeling for task 1, the training process optimizes the model to 1) given an input image in class $c$ for task 1 (MNIST), minimize the value observed at $(c+1)^{th}$ detector, as well as maximize the values observed at other detectors; 2) given an input image in class $c$ for task 1 (Fashion-MNIST), maximize the value observed at $(c+1)^{th}$ detector, as well as minimize the values observed at other detectors. Thus, the resulting multi-task model is able to automatically differentiate which task the input image belongs to based on the sum of values observed in the ten detectors, and then generate the predicted class using \texttt{argmin} (\texttt{argmax}) function for MNIST (Fashion-MNIST) task. The gradient updates have been summerized in Equation \ref{eq:graident_updates}. 

\begin{equation}
\begin{multlined}
\theta^{share} = \theta^{share} - \frac{1}{2} \cdot \eta \frac{\lambda_2}{\lambda_1}(\nabla \theta^1 + \nabla \theta^2)\\
\theta^{1'} =  \theta^{1} - \eta \nabla \theta^1 - 2\eta \lambda_{L2}|| \theta^{1} + \theta^{2}||\\
\theta^{2'} =  \theta^{2} - \eta \nabla \theta^2 - 2\eta \lambda_{L2}|| \theta^{1} + \theta^{2}||\\
     \end{multlined}
     \label{eq:graident_updates}
\end{equation}

\paragraph{System Noise Modeling} We demonstrate that the proposed system is robust under the noise impacts from the device variations of diffractive layers and the detector noise in our system. Specifically, to include the noise attached to the detector, we generate a Gaussian noise mask $\mathcal{N}(\sigma, \mu) \in \mathbb{R}^{200\times200}$ with on the top of the detector readings, i.e., each pixel observed at the detector will include a random Gaussian noise. As shown in Figure \ref{fig:noise_eval}(a), we evaluate our system under multiple Gaussian noises defined with different $\sigma$ with $\mu=0$. We also evaluated the impacts of $\mu$, while we do not observe any noticeable effects on the accuracy for both tasks. This is because increasing $\mu$ of a Gaussian noise tensor does not change the ranking of the values observed by the ten detectors, such that it has no effect on the finalized classes generated with \texttt{argmax} or \texttt{argmin}. The forward function for $i^{th}$ task with detector noise is shown in Equation \ref{eq:det_noise}. 

\begin{equation}
    c^i = \texttt{argmax/argmin}(det(f(\theta^{share}, \theta^i, \mathcal{X}^i)) + \mathcal{N}(\sigma, 0)), ~~i=\{1,2\}
    \label{eq:det_noise}
\end{equation}

We also considered the imperfection of the devices used in the system. With 3D printing or lithography based techniques, the imperfection devices might not implement exactly the phase parameters optimized by the training process. Specifically, we consider the imperfection of the devices that affect the phases randomly under a Gaussian noise. As shown in Figure \ref{fig:noise_eval}(b), the x-axis shows that the $\sigma$ of Gaussian noise that are added to the phase parameters for inference testing. The forward function is described in Equation \ref{eq:phase_noise}. Beam splitter noise has also been quantified, where we do not see direct impacts on both tasks (see Figure 2 in supplementary file SI.pdf). 

\begin{equation}
\begin{multlined}
    \theta^{share}_{\mathcal{N}} = (\theta^{share} + \mathcal{N}(\sigma, 0)) ~~\% ~~2\pi\\
    \theta^{i}_{\mathcal{N}} = (\theta^{i} + \mathcal{N}(\sigma, 0)) ~~\% ~~2\pi, ~~i=\{1,2\}\\
    c^i = \texttt{argmax/argmin}(det(f( \theta^{share}_{\mathcal{N}}, \theta^i_{\mathcal{N}}, \mathcal{X}^i_{\mathcal{N}}))), ~~i=\{1,2\} \\
   \end{multlined}
    \label{eq:phase_noise}
\end{equation}

Finally, for results shown in Figure \ref{fig:noise_eval}(c)--(d), we include both detector noise and device variations in our forward function (Equation \ref{eq:combined_noise}):

\begin{equation}
\begin{multlined}
    \theta^{share}_{\mathcal{N}} = (\theta^{share} + \mathcal{N}^1(\sigma^1, 0)) ~~\% ~~2\pi\\
    \theta^{i}_{\mathcal{N}} = (\theta^{i} + \mathcal{N}^1(\sigma^1, 0)) ~~\% ~~2\pi, ~~i=\{1,2\}\\
    c^i = \texttt{argmax/argmin}(det(f( \theta^{share}_{\mathcal{N}}, \theta^i_{\mathcal{N}}, \mathcal{X}^i_{\mathcal{N}})) + \mathcal{N}^2(\sigma^2, 0))), ~~i=\{1,2\} \\
   \end{multlined}
    \label{eq:combined_noise}
\end{equation}



\bibliography{bibs/dlsys_yu.bib,bibs/analog.bib}
\bibliographystyle{Science}

\section*{Acknowledgments}
Y.C. thanks the support from grants NSF-2019336 and NSF-2008144. Y.C. and W.G. thank the support from the University of Utah start-up fund. Y.C. thanks the support from grants NSF-2019336 and NSF-2008144.

\clearpage

\section*{Supplementary materials}

\paragraph{Derivation of D$^2$NN} The forward propagation follows the model described in [17], where the light propagation consists of free space propagation of diffractive light and transparency phase modulation, as shown in Figure 6 (in the main file). The phase modulation follows the direct multiplication of input wavefunction and phase function of masks, and free space propagation features the interconnects (addition of wavefunctions) between layers. Here, we details the treatment of free space propagation implemented in our model. 

In experimental process, as shown in Figure 6 (in the main file), the input light passes through a set of diffractive layers and then is collected by output detectors. When light is incident on one diffractive layer, according to \textit{Huygens-Fresnel} principle , each point on the output of the diffractive layer can be seen as a secondary point source emitting spherical waves and the output wave is the sum of all these spherical waves. It means the inputs of every point on $l$-th layer can be seen as the sum of outputs from all the points on $(l-1)$th layer. This features the interconnects in a neural network and each point acts as a neuron.

The free-space propagation relation is
\begin{equation}
g(x,y)=f(x,y)h(x,y),
\end{equation}
where $f(x,y)$ is input function and $h(x,y)$ is the impulse response function of free space. 
Under Fresnel's approximation, the impulse response function is:
\begin{equation}
h(x,y){\approx}\frac{j}{{\lambda}z}\text{exp}(-jkz)\text{exp}(-j{\pi}\frac{x^2+y^2}{{\lambda}z})
\end{equation}
Thus, the input at point $(x,y)$ on $l$-th layer can be written as the sum of all the outputs at $(l-1)$-th layer:
\begin{equation} \label{freespacep}
g_l(x,y)={\iint}f_{l-1}(x',y')h(x-x',y-y')dx'dy' , 
\end{equation}
where
\begin{equation}
h(x-x',y-y'){\approx}\frac{j}{{\lambda}d}\text{exp}(-jkd)\text{exp}(-j{\pi}\frac{(x-x')^2+(y-y')^2}{{\lambda}d})
\end{equation}
and $d$ is the distance between layers. $f_{l-1}$ is the output wavefunction of points on $(l-1)$th plane and also the input wavefunction of free space propagation, $g_l$ is the output function of free-space propagation and also input for the phase mask at $l$-th plane.

In simulation process, the convolution of Eq.\,\ref{freespacep} is complicated for implementation and training. To speed up and simplify the training process, especially under \texttt{Pytorch} framework, Fourier transform to spatial frequency domain is employed. 
Specifically, the free space propagation output wavefunction (\ref{freespacep}) is in convolution form $f_{l-1}*h$. By convolution theorem, the Fourier transform of the convolution is the product of Fourier transforms of $f_{1-1}$ and $h$:

\begin{equation}
\mathcal{F}(g_l(x,y))=\mathcal{F}(f_{l-1}(x,y))\mathcal{F}(h(x,y)) 
\end{equation}

\begin{equation}\label{eq_iofourier}
    U_l(f_x,f_y)=F_{l-1}(f_x,f_y)H(f_x,f_y)
\end{equation}

The Fourier transform of $h$ is
\begin{equation}
     H(f_x,f_y)=\mathcal{F}(h(x,y))=\text{exp}(jkd){\text{exp}}(j{\pi\lambda}z(f_x+f_y)^2)
\end{equation}
This is the free space transfer function.

\begin{figure} 
    \centering
    \includegraphics[scale=0.6]{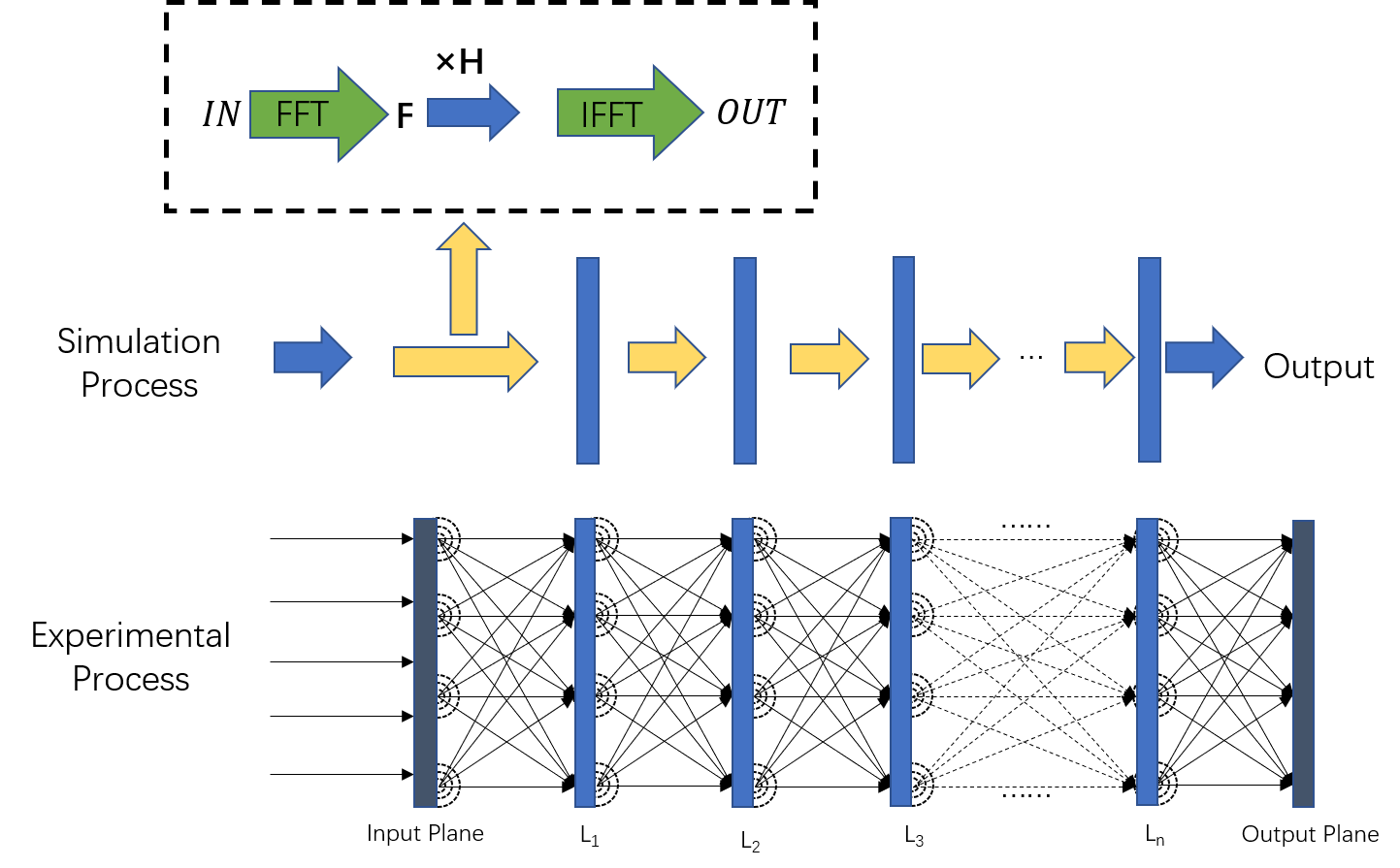}
    \caption{Simulation and experimental process}
    \label{fig_ex-sim}
\end{figure}

As shown in Figure \ref{fig_ex-sim}, for the free space propagation between layers, the signal is firstly converted to spatial frequency domain ($f_x, f_y$) through fast Fourier transformation (FFT). According to Equation (\ref{eq_iofourier}), the output of free space propagation is simply the product of the input FFT signal and free space transfer function $H$. After the free space propagation, the FFT signal is converted back to spatial domain ($x,y$) through inverse Fourier transformation (IFFT) and the obtained signal is modulated through the phase mask. 

\begin{figure}[!htb]
       \centering
      \includegraphics[width=1\linewidth]{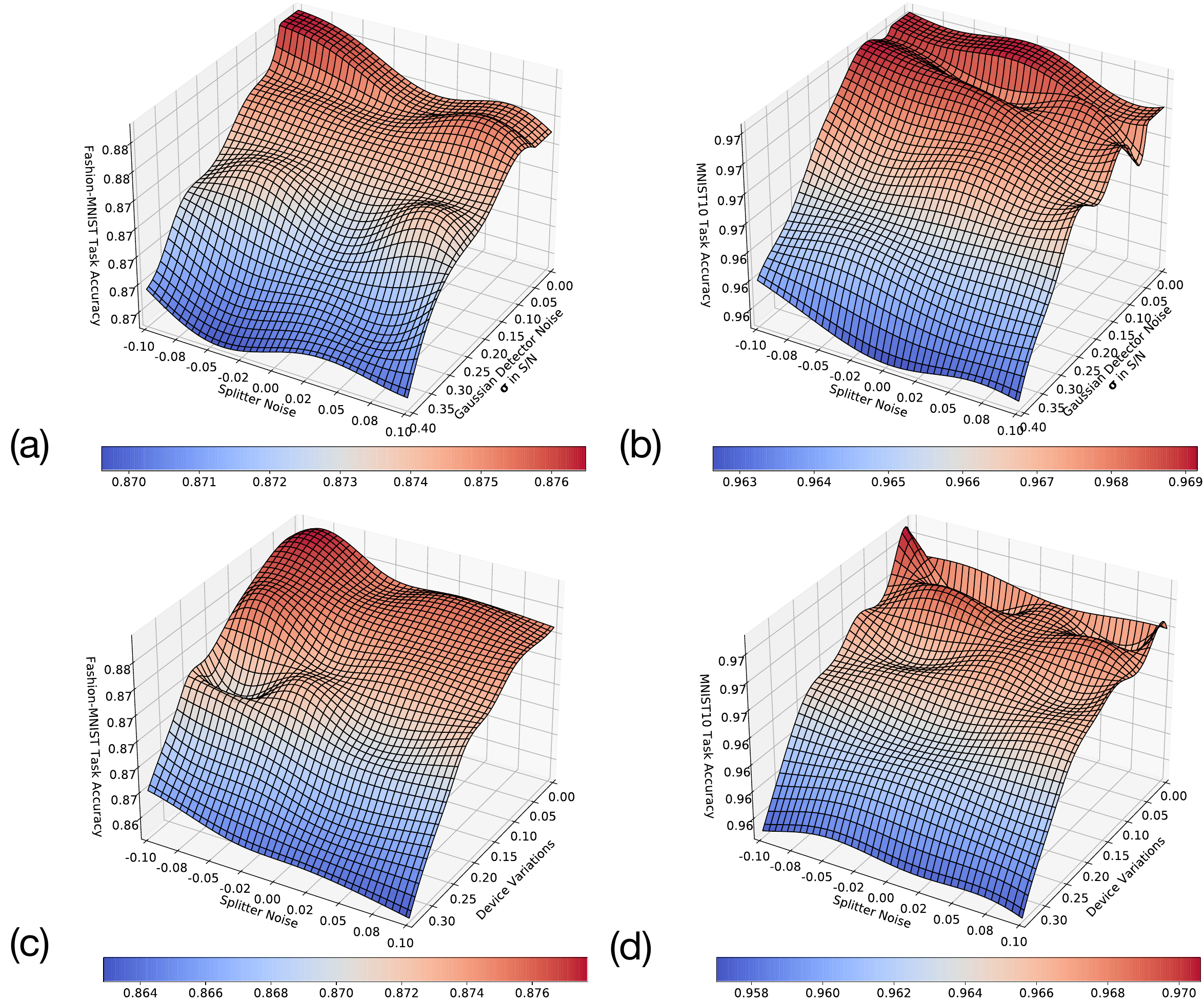}
            \caption{{\bf Evaluation of 50-50 beam splitter noise with device variations and detector noise.} (a)--(b) Prediction performance evaluation under Gaussian detector noise with $\sigma$ shown in $S/N$ (Signal to Noise) $\in [0, 0.2]$ and beam splitter noise $\in [-0.1, 0.1]$, with accuracy of MNIST shown in (a) and accuracy of Fashion-MNIST shown in (b). (c)--(d) Prediction performance evaluation under Gaussian device variation noise with $\sigma \in [0,0.15]$,  and beam splitter noise $\in [-0.1, 0.1]$, with accuracy of MNIST shown in (c) and accuracy of Fashion-MNIST shown in (d).}
    \label{fig:splitter_noise}
\end{figure}

\paragraph{Beam splitter noise evaluation} Similar to noise evaluation shown in Figure 4 (in the main file), we evaluate the prediction performance under two different noises, i.e., beam splitter noise with detector noise and beam splitter noise with device variations. The noise modeling has been discussed in Section Methods. Figure \ref{fig:splitter_noise} (in this SI file) present the prediction accuracy of both tasks under these two noise set ups. We can see that the splitter noise does not have significant impacts on the accuracy compared to other system noise. As shown in Figure \ref{fig:splitter_noise}(a)--(b) (in this SI file), the accuracy degradation is basically dominated by the detector noise; and as shown in Figure \ref{fig:splitter_noise}(c)--(d) (in this SI file), the accuracy degradation is basically dominated by the device variation.

{\color{black}
\paragraph{Multi-task Architecture Beyond Two Tasks} The proposed system design methodology can be extended to deal with multi-task DNN tasks with more than two tasks. Here, we describe the implementations of how it can be extended a 4-task D$^2$NNs, and the experimental results of 4-tasks D$^2$NNs evaluated using MNIST, Fashion-MNIST, K-MINIST, and E-MNIST are included as well. First, let?s review the experimental setup in the initial submitted version. In the initial model setup, four common layers and two separated layers for each channel (two channels in total for two tasks). At the output of the model, each detector is reading 20$\times$20 pixel value of the pre-defined regions, and ten detector regions are employed to encode 10 class using a positive one hot encoding and a reversed one hot encoding. Then, typical one-hot encoding and the reverse one-hot encoding are used to separate two different tasks with ten detector regions, such that it successfully encodes two datasets/tasks simultaneously.

To implement 4 different tasks without minimum $HWCost$ overhead (i.e., minimizing the number of detectors needed), we implement a 10-detector version for the 4-task D$^2$NNs, where each detector region is encoded with two digits (see Figure \ref{fig:4tasks}). For one detector region, it is split into two equal parts, one part for Task 1 and the other one for Task 2. In this way, ten detector regions with each size of 20$\times$20 are expanded to twenty detector regions with each size of 20*10. In which case, we are using the same number of detectors as proposed in the submitted paper. Here, using label \texttt{9} for each task as an example, the extended label modeling of four different tasks are illustrated below:
\begin{itemize}
\item for the first task is read as  $[0, 0, 0, 0, 0, 0, 0, 0, 0, 0, 0, 0 ,0, 0, 0, 0, 0, 0 ,1, 0]$, 
\item for the second task is read as $[0, 0, 0, 0, 0, 0, 0, 0, 0, 0, 0, 0 ,0, 0, 0, 0, 0, 0 ,0, 1]$, 
\item for the third task is read as    $[1, 1, 1, 1, 1, 1, 1, 1, 1, 1, 1, 1 ,1, 1, 1, 1, 1, 1 ,1, 0]$, 
\item for the forth task is read as     $[1, 1, 1, 1, 1, 1, 1, 1, 1, 1, 1, 1 ,1, 1, 1, 1, 1, 1 ,0, 1]$. 
\end{itemize}

For aforementioned example, the pattern shown on the detector will be like the following (Figure \ref{fig:4tasks}):
 
\begin{figure}[!htb]
       \centering
      \includegraphics[width=1\linewidth]{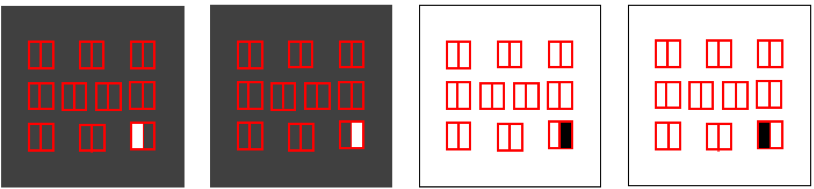}
            \caption{{\bf Illustration of detector patterns for class \texttt{9} of four different image classification tasks.}}
    \label{fig:4tasks}
\end{figure}

\textbf{Experimental results of 4-Task D$^2$NNs}: We implement the aforementioned 4-Task D$^2$NNs using the same PyTorch-based framework. Regarding the four tasks, MNIST is used as the first task, Fashion-MNIST (F-MNIST) is used as second task, K-MNIST (Kuzushiji letters) is used as the third task and the first ten classes in E-MNIST (English letters) is used as the fourth task. All these four datasets have ten labels to classify. To make it comparable to the initial model setup, we have four common layers and two separate layers for each channel (four channels are implemented, each channel for one task). Thus, the total diffractive layers of the 4-Task system is 12, where the number of detectors remains the same. The training setups remain the same as the initial implementation for the 2-Task model. The details of accuracy, cost of the system, and cost efficiency value Acc-HW product are included in Table \ref{tbl:4tasks}.

\begin{table}[!htb]
\caption{Evaluation results of 4-Task D2NNs based on the proposed approach, and comparisons with baseline Single-task D2NNs models}
\label{tbl:4tasks}
\begin{tabular}{|l|l|l|l|l|l|l|}
\\
 & \textit{\textbf{\# Diffractive Layers}} & \textit{\textbf{\# Detectors}} & \textit{\textbf{MNIST}} & \textit{\textbf{F-MNIST}} & \textit{\textbf{K-MNIST}} & \textit{\textbf{E-MNIST}} \\ \hline
\textbf{\begin{tabular}[c]{@{}l@{}}4-Task \\ D2NNs\end{tabular}} & 4 + 2$\times$4 = 12 & 10 & 0.958 & 0.857 & 0.860 & 0.884 \\ \hline
\textbf{\begin{tabular}[c]{@{}l@{}}Single-Task \\ D2NNs\end{tabular}} & 5$\times$4 = 20 & 10$\times$4 = 40 & 0.981 & 0.889 & 0.861 & 0.909 \\ \hline
\textbf{\begin{tabular}[c]{@{}l@{}}Acc-HW Prod\end{tabular}} & - & - & 3.91$\times$ & 3.85$\times$ & 4.00$\times$ & 3.89$\times$ \\ \hline
\end{tabular}
\end{table}

In Table \ref{tbl:4tasks}, we can see that our setup can be easily extended to deal with multiple tasks since we can split the detector region with size 20$\times$20 to fit the number of tasks. While we observe that the accuracy of each task is degraded due to the 4-Task labeling rules, the cost efficiency of the system is further increased by approximately 2$\times$ compared to our 2-Task model (3.85$\times$--4$\times$).

However, there are potential degradations in cost efficiency as we increase the number of tasks. As we increase the number of tasks implemented with our model, the detector region for each class in each task will be smaller, which means the result may be easily affected by noise and the accuracy can be degraded. Some accuracy and cost efficiency degradations are expected compensated when we increase the number of tasks of the model by limiting the number of detectors. Specifically, in this work, we focus on cost efficiency optimization. Note that cost efficiency is related to two metrics: 1) accuracy and 2) HW cost. Thus, while targeting more tasks in the proposed system (e.g., 8 or 16 tasks), a detailed trade-off analysis will be needed to find the optimal cost efficiency.  
}

\begin{figure}[!htb]
       \centering
      \includegraphics[width=0.5\linewidth]{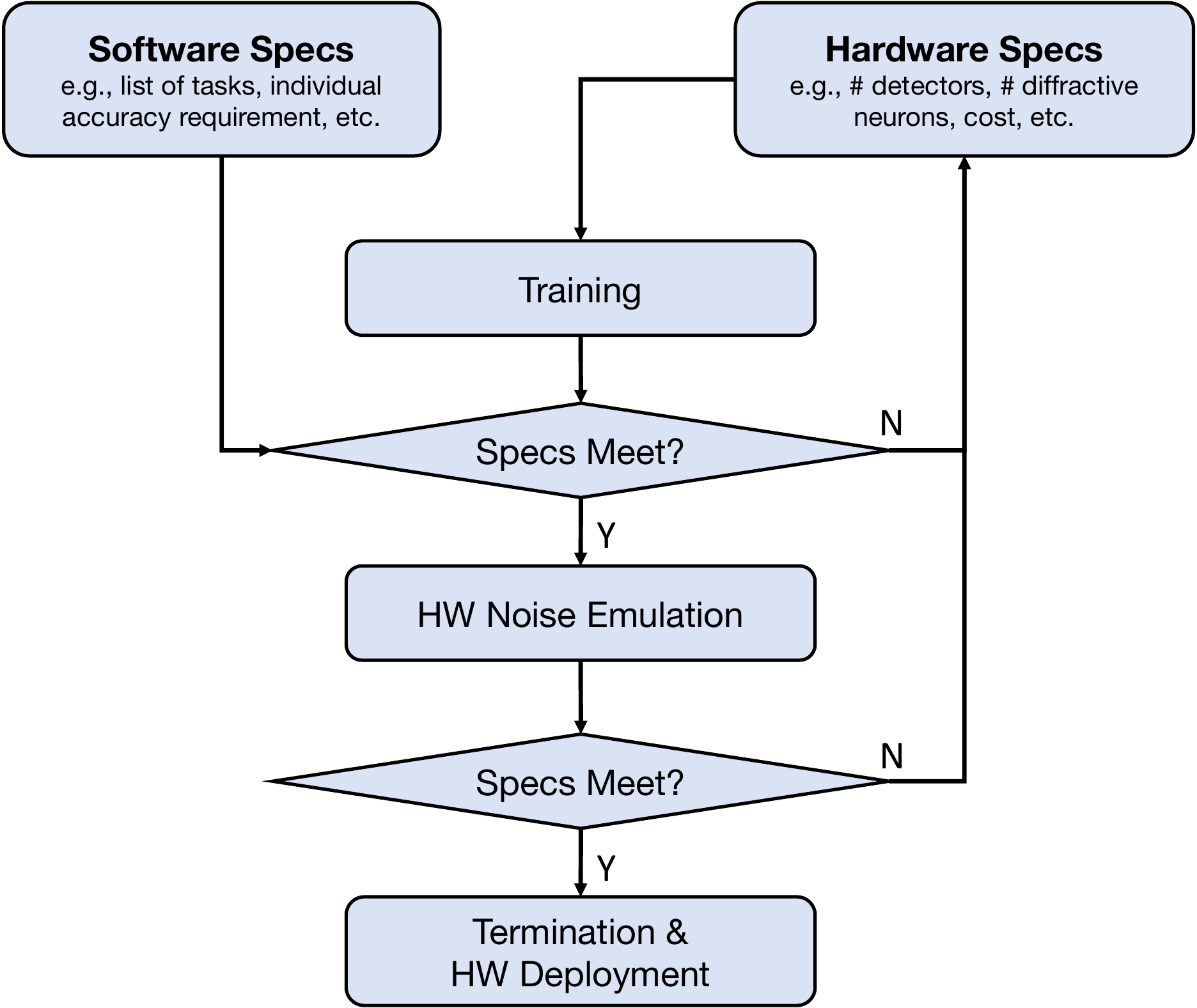}
            \caption{{\bf Illustration of hardware-software co-design flow for multi-task D$^2$NNs}}
    \label{fig:codesign-flow}
\end{figure}

{\color{black}
\paragraph{Hardware-software co-design flow} The design flow of proposed work incorporates iterative design process by 1) adjusting the machine learning objectives, and 2) the expected hardware cost, shown in Figure \ref{fig:codesign-flow}. Specifically, with the proposed training methods that is guided by adjustable learning strengths of each individual tasks, the proposed design method will first optimize the network parameters for task-specific multi-task learning objectives. While at this stage, the initial neural architecture has been developed and will be refined based on the noise-free model evaluation results. Once the learning objectives are satisfied, the estimated or measured system noise will be elaborated into the inference model. Similarly, the system will be optimized iteratively in co-design fashion by adjusting the training and hardware setups. Finally, the model parameters will be generated and deployed to specific hardware devices (e.g., 3D printing or SLMs).  We have updated the supplementary information with the flow explanations. }

\clearpage

\end{document}